\pdfoutput=1

\documentclass[11pt]{article}

\usepackage{naacl2021}

\usepackage{times}
\usepackage{latexsym}

\usepackage[T1]{fontenc}

\usepackage[utf8]{inputenc}

\usepackage{microtype}

\usepackage[inline]{enumitem}
\usepackage[pdftex]{graphicx}
\usepackage{multirow}
\usepackage{breqn}
\usepackage{amsmath}
\usepackage[caption=false]{subfig}
\usepackage{amsmath,amsfonts,amssymb}

\title{UmlsBERT: Clinical Domain Knowledge Augmentation of Contextual Embeddings Using the Unified Medical Language System Metathesaurus}

\author{%
George Michalopoulos, Yuanxin Wang, Hussam Kaka, Helen Chen, Alexander Wong \\
    University of Waterloo, \\  Waterloo, Canada \\
\{gmichalo, yuanxin.wang, hussam.kaka, helen.chen, alexander.wong\}@uwaterloo.ca
}

\begin{document}
\maketitle
\begin{abstract}
Contextual word embedding models, such as BioBERT and Bio\_ClinicalBERT, have achieved state-of-the-art results in biomedical natural language processing tasks by focusing their pre-training process on domain-specific corpora.  However, such  models do not take into consideration structured expert domain knowledge from a knowledge base. 

We introduce UmlsBERT, a contextual embedding model that integrates domain knowledge during the pre-training process via a novel knowledge augmentation strategy.  More specifically, the augmentation on UmlsBERT with the Unified Medical Language System (UMLS) Metathesaurus is performed in two ways: \begin{enumerate*}[label=(\roman*)] \item connecting words that have the same underlying `concept' in UMLS and \item leveraging semantic type knowledge in  UMLS to create clinically meaningful input embeddings.\end{enumerate*} By applying these two strategies,
UmlsBERT can encode clinical domain knowledge into word embeddings and outperform existing domain-specific models on common named-entity recognition (NER) and clinical natural language inference  tasks.
\end{abstract}

\section{Introduction}
In recent years, the volume of data being collected in healthcare has grown considerably. A significant proportion of the data is in text form, which requires advanced Natural Language Processing (NLP) models to process.  This has led  to the creation of high-performing, optimized NLP models focused on the biomedical domain.

 Contextual word  embedding  models, such  as  ELMo   \citep{peters-etal-2018-deep} and BERT \citep{Devlin2019BERTPO} have achieved state-of-the-art results in many NLP tasks.  Initially tested in a general domain, these models have also been successfully applied in the biomedical domain by pre-training them on biomedical corpora, leading to the best performances in  a variety of biomedical NLP tasks \citep{10.1093/bioinformatics/btz682},  \citep{alsentzer-etal-2019-publicly}. However, current biomedical applications of transformer-based Natural Language Understanding (NLU) models do not incorporate 
 structured expert domain knowledge from a knowledge base
 into their embedding pre-training process.

The Unified Medical Language System (UMLS) \cite{umls} Metathesaurus is a compendium of many biomedical terminologies with the associated information, such as synonyms and categorical groupings. It allows for the connection of words that represent the same or similar `concept'. For example, the words `lungs' and `pulmonary' share a similar meaning and thus can be mapped to the same  concept unique identifier (CUI) \textit{CUI: C0024109}. Additionally,  UMLS allows the grouping of concepts according to their semantic type \cite{Aggarticle}. For example, `skeleton' and `skin' have the same `Body System' semantic type, and `inflammation' and `bleed' are in the `Pathologic Function' semantic type.

In this paper, 
we present and publicly release\footnote{https://github.com/gmichalo/UmlsBERT} a novel architecture for augmenting contextual embeddings with clinical domain knowledge. Specifically:
\begin{enumerate*}[label=(\roman*)] 
\item We are the first, to the best of our knowledge, to propose the  usage of domain (clinical) knowledge from a clinical Metathesaurus (UMLS Metathesaurus) in  the pre-training phase of a BERT-based model (UmlsBERT) in order to build `semantically enriched' contextual representations that will benefit from  both the contextual learning (BERT architecture)  and the domain knowledge (UMLS Metathesaurus).
\item We propose a new  multi-label loss function for the pre-training of the  Masked Language Modelling (Masked LM) task in the UmlsBERT that incorporates the connections between clinical words using the CUI attribute of UMLS.  
\item  We introduce a  semantic type embedding  that enriches the input  embeddings process of the UmlsBERT  by forcing the model  to take into consideration the association between words that are of the same semantic type.
\item Finally, we demonstrate that UmlsBERT outperforms two popular clinical-based BERT models (BioBERT and Bio\_ClinicalBERT) and a general domain BERT model  on   different clinical named-entity recognition (NER) tasks and on one clinical natural language inference task.
\end{enumerate*}

The rest of paper is organized as follows. Related work is presented in Section \ref{related}. The data that were used to pre-train and test the new UmlsBERT are described in Section \ref{data}. The characteristics of the proposed UmlsBERT architecture for augmenting contextual embeddings with clinical knowledge are detailed in Section \ref{methods}. Finally, the results of the down-stream tasks and the qualitative analysis are reported in Section \ref{results},  and a conclusion and a plan for future work are presented in Section \ref{conlusion}.

\section{Related Work}
\label{related}

In \citep{peters-etal-2018-deep}, contextualized word embeddings were introduced in a bidirectional language model (ELMo). This allowed the model to change the embedding of a word based on its imputed meaning, which was derived from the surrounding context. 
Subsequently, \citep{Devlin2019BERTPO} proposed the  Bidirectional  Encoder  Representations  from  Transformers  (BERT)  which used bidirectional transformers \citep{Vaswani2017AttentionIA} to create context-dependent representations. 
For both models, pre-training is done  on massive corpora and the context-sensitive embeddings can be used for  downstream tasks. 

Other approaches enhance the BERT's performance by injecting external knowledge from a knowledge base. Sense-BERT \citep{Levine2020SenseBERTDS}  is pre-trained to predict the supersenses (semantic class) of each word by incorporating lexical semantics (from the lexical database WordNet \citep{wordnet}) into the model's pre-training objective and by adding supersense information to the input embedding. In addition, GlossBERT \citep{glossbert} focuses on improving word sense disambiguation by using context-gloss pairs on the 
sentence-pair classification task of a BERT model.

Furthermore, there have been multiple attempts to improve the performance of contextual models in the biomedical domain. BioBERT is a BERT-based model which was pre-trained on both general (BooksCorpus and English Wikipedia) and biomedical corpora (PubMed abstracts and PubMed Central full-text articles) \citep{10.1093/bioinformatics/btz682}. The authors demonstrate that incorporating biomedical corpora in the pre-training process improves the performance of the model in downstream biomedical tasks. This is 
likely  because medical corpora contains   terms  that are not usually found in a general domain corpus \citep{10.1093/bioinformatics/btx228}.
Finally, Bio\_ClinicalBERT \citep{alsentzer-etal-2019-publicly} further pre-trains BioBERT on clinical text from the MIMIC-III v1.4 database \citep{mimiciii}. It is shown that the usage of clinical specific contextual embeddings  can be beneficial for the performance of a model on different clinical NLP down-stream tasks.

\section{Data}
\label{data}
We use the Multiparameter  Intelligent  Monitoring  in  Intensive  Care III (MIMIC-III) dataset \citep{mimiciii} to pre-train the UmlsBERT model. MIMIC dataset consists of anonymized electronic medical records in English of over forty-thousand patients who were admitted to the intensive care units of the Beth Israel Deaconess Medical Center (Boston, MA, USA) between 2001 and 2012. In particular, UmlsBERT  is trained on the \textbf{NOTEEVENTS} table, which contains 2,083,180 rows of clinical notes and test reports.

 \begin{table}[h]
 \centering
  \begin{tabular}{c c c c c}
  \hline
  
Dataset &  Train  &   Dev  &  Test & C     \\  \hline

MedNLi & 11232 & 1395 & 14 22 & 3 \\
i2b2 2006 & 44392 &  5547& 18095 & 17 \\
i2b2  2010 & 14504 &  1809 & 27624 & 7 \\
i2b2 2012 & 6624 & 820 & 5664 & 13 \\
i2b2 2014 & 45232 & 5648 & 32586 & 43 \\
  \end{tabular}
  \caption{Number of sentences for the train/dev/test set of each dataset. We also include the number of classes (C) for each dataset. We use the same splits that are used in \citep{alsentzer-etal-2019-publicly}.}
  \label{tab:dataset}
\end{table}

We evaluate the effects of the novel features of the UmlsBERT model  on the English MedNLI natural language inference task \citep{MEDNLIinproceedings} and on four i2b2 NER tasks (in IOB format \citep{iob}). More specifically, we experiment on the following  English i2b2 tasks:  the i2b2  2006 de-identification challenge \citep{article2006},  the i2b2 2010 concept extraction challenge \citep{2010article}, the i2b2 2012 entity extraction challenge \citep{2010article} and the i2b2 2014 de-identification  challenge \citep{Stubbs2015AutomatedSF}. 
These datasets are chosen because of their use in benchmarking  prior biomedical BERT models, thereby allowing for performance comparison. In addition, these publicly available datasets  enable the reproducibility of our results and meaningful comparison with future studies. Table \ref{tab:dataset} lists the statistics of all the datasets. Finally, it should be noted that for the identification of the UMLS terms, we use the UMLS 2020AA version.

\section{Methods}
\label{methods}


\begin{figure*}[h]
\centering
    \includegraphics[height=3.8cm]{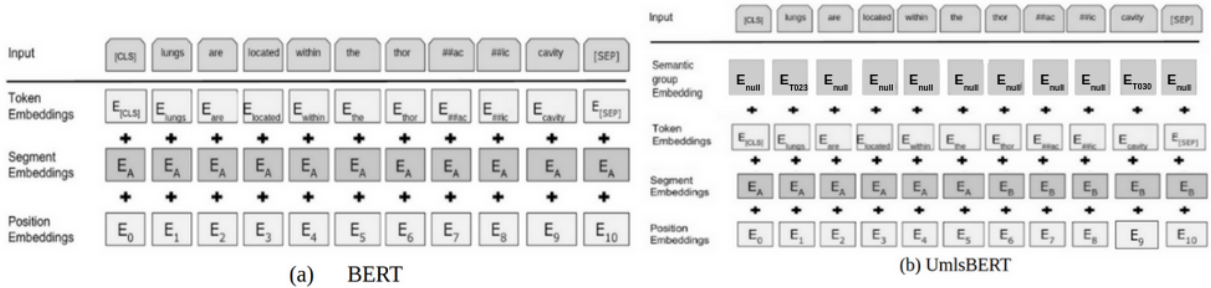}

  \caption{  \textbf{(a)} Original  input vector of the  BERT model \citep{Devlin2019BERTPO}. \textbf{(b)} Augmented input vector of the UmlsBERT where the semantic type embeddings is available. For the words `lungs' and `cavity', their word embeddings are enhanced with the embedding of the semantic type `Body Part, Organ, or Organ Component'($E_{T023}$) and `Body Space or Junction'($E_{T030}$) respectively. The rest of the words are not related to a medical term, so a zero-filled tensor $E_{null}$ is used. }
  \label{fig:semb1}
\end{figure*}

\subsection{BERT Model}
\label{sec:bert}  
 The original BERT model \citep{Devlin2019BERTPO} is  based on multi-layer bidirectional transformers \citep{Vaswani2017AttentionIA}, which generates contextualized word representations. Incorporating information from bidirectional representations allows the BERT model to  capture more accurately the meaning of a word based on its surrounding context, i.e. sentence.
 
The pre-training phase of the BERT model consists of two self-supervised tasks: Masked Language Modelling (LM), in which a percentage of the input is masked at random and the model is forced to predict the masked tokens, and Next Sentence Prediction, in which the model has to determine whether two segments appear consecutively in the original text. Since our UmlsBERT model is focused on augmenting the Masked LM task with clinical information from the UMLS Metathesaurus, we omit the description of the Next Sentence Prediction task and only describe the details of the Masked LM task herein.

In Masked LM, $15\%$ of the tokens of each sentence are replaced by a [MASK] token. For the $j^{th}$ input token in the sentence, an input embedding vector $u^{(j)}_{input}$ is created by the following equation:

\begin{equation}
u^{(j)}_{input} = p^{(j)} + SEG seg_{id}^{(j)} + Ew_j
\label{eq:1}
\end{equation}

where $p^{(j)} \in \mathbb{R}^{d}$ is the position embedding of the $j^{th}$ token in the sentence, and $d$ is the transformer's hidden dimension. Additionally, $SEG \in \mathbb{R}^{ d \times 2}$ is called  the segment embedding, and  $seg_{id} \in \mathbb{R}^{2}$, a 1-hot vector, is the segment id that indicates the sentence to which the token belongs. In Masked LM, the model uses only one sentence and therefore, the segment id indicates that all the tokens belong to the first sentence. $E \in \mathbb{R}^{d \times D}$ 
is the token embedding where $D$ is the length of the model's vocabulary and $w_j \in \mathbb{R}^{D}$ is a 1-hot vector corresponding to the $j^{th}$ input token. 

The input embedding vectors pass through multiple attention-based transformer layers  where each layer produces a contextualized embedding of each token. Finally, for each masked token $w$, the model outputs a score vector  $y_w   \in \mathbb{R}^{D} $ with the goal of minimizing the cross-entropy loss between the softmax of  $y_w $ and the 1-hot vector corresponding to the masked token $(h_w)$:
\begin{equation}
\label{loss1}
    loss = -log(\frac{exp(y_w [w])}{\sum_{w'} exp(y_w [w'])})
\end{equation}

\subsection{Enhancing Contextual Embeddings with Clinical Knowledge}
\label{sec:clkn}
In the UmlsBERT model, we update the Masked LM procedure to take into consideration the associations between the words specified in the UMLS Metathesaurus.

\subsubsection{Semantic type embeddings}
We introduce a new embedding matrix called $ST \in  \mathbb{R}^{ D_s \times d}$ into the input embedding of the BERT model, where $d$ is BERT's transformer hidden dimension and $D_s=44$ is the number of unique UMLS semantic types that can be identified in the vocabulary of our model. In particular, in this matrix, each row represents the  unique semantic type in UMLS that a word can be identified with (for example the word `heart' is associated with the semantic type T023:`Body Part, Organ, or Organ Component' in UMLS). 

To incorporate the $ST$ embedding matrix into the input embedding of our model, all words with a clinical meaning defined in UMLS are identified. For each of these words, the corresponding concept unique identifier (CUI) and  semantic type  are extracted. We use $s_w \in \mathbb{R}^{D_s} $ as a 1-hot vector corresponding to the semantic type of the medical word $w$. The identification of the UMLS terms and their UMLS semantic type is accomplished using the open-source Apache clinical Text Analysis and Knowledge Extraction System (cTakes) \citep{ctakesarticle}. Thus, by introducing the semantic type embedding, the input vector (equation \ref{eq:1}) for each word is updated to: 

\begin{equation}
u^{(j)\prime}_{input} = u^{(j)}_{input} + ST^\top s_w
\label{eq:3}
\end{equation}

where the semantic type vector $ST^\top s_w$ is set to a zero-filled vector for words that are not identified in UMLS. 

We hypothesize that incorporating the clinical information of the semantic types into the input tensor could be beneficial for the performance of the model as 
the semantic type representation  can be used to enrich the input vector of words that are rare in the training corpus and the model do not have the chance to learn meaningful information for their representation. 
Figure \ref{fig:semb1} presents an overview of the insertion of the semantic type embeddings into the standard BERT architecture.

\subsubsection{Updating the loss function of Masked LM task}

Furthermore, we update the loss function of the Masked LM pre-training task to take into consideration the connection between words that share the same CUI. As described in Subsection \ref{sec:bert}, the loss function of the Masked LM pre-training task of a BERT model is a cross-entropy loss between the softmax vector of the masked word and  the 1-hot vector that indicates the actual masked word. We proposed to `soften' the loss function and updated it to a multi-label scenario by using information from the CUIs.

More specifically, instead of using a 1-hot vector $(h_w)$ that corresponds only to the masked word $w$, we use a binary vector indicating the presence of  all the words which shared the same CUI of the masked word $(h_w^\prime)$. Finally, in order for the model to properly function in a multi-label scenario, the cross entropy loss (equation \ref{loss1})  is updated to a binary cross entropy loss:

\begin{multline}
    \label{loss2}
loss = \sum_{i=0} ^D (- h_w^{'}[i] log(y_w [i])  \\  + (1- h_w^{'}[i]) log(1- y_w [i]))
\end{multline}

These changes force UmlsBERT to learn the  
semantic relations between words, which  are associated with the same CUI in a biomedical context.

\begin{figure}[h]
   \centering
 \includegraphics[width=0.451\textwidth]{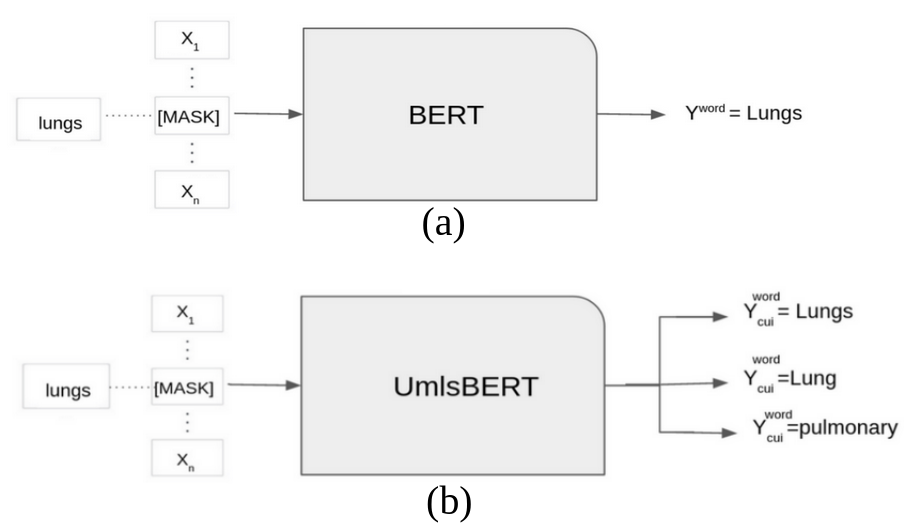}
  \caption{An example of predicting the masked word `lungs' \textbf{(a)} the BERT model tries to predict only the word lungs \textbf{(b)} whereas the UmlsBERT tries to identify  all words that are associated with the same CUI (e.g lungs, lung, pulmonary).}
  \label{fig:semb}
\end{figure}

 An example of predicting the masked word `lungs' with and without the clinical information is presented in Figure \ref{fig:semb}. As seen in this figure, the UmlsBERT model tries to identify the words `lung', `lungs' and `pulmonary' because all three words are associated with the same \textit{CUI: C0024109} in the UMLS Metathesaurus.
 
 \begin{table*}
 \centering
  \begin{tabular}{l c c c c c}
  \hline

Dataset &   & BERT\textsubscript{based}  &     BioBERT &   Bio\_ClinicalBERT &    UmlsBERT   \\ \hline
\multirow{3}{*}{MedNLI}  &  epochs& $4$ & $4$ & $4$ & $3$   \\  
 &  batch size  & $16$  & $16$  & $32$  & $16$     \\
  &  learning rate  & $5\mathrm{e}$-$5$    & $3\mathrm{e}$-$5$    &$3\mathrm{e}$-$5$    &$3\mathrm{e}$-$5$        \\ 
 \hline
 
 \multirow{3}{*}{i2b2 2006}  &    epochs& $20$ & $20$ & $20$ & $20$   \\  
 &  batch size  & $32$  & $16$  & $16$  & $32$     \\
  &  learning rate  & $2\mathrm{e}$-$5$    & $2\mathrm{e}$-$5$    &$2\mathrm{e}$-$5$    &$5\mathrm{e}$-$5$        \\  \hline

 \multirow{3}{*}{i2b2 2010}  &  epochs& $20$ & $20$ & $20$ & $20$     \\  
 &  batch size  & $16$  & $32$  & $32$  & $16$     \\
  &  learning rate  & $3\mathrm{e}$-$5$    & $3\mathrm{e}$-$5$    &$5\mathrm{e}$-$5$    &$5\mathrm{e}$-$5$        \\  \hline

 \multirow{3}{*}{i2b2 2012}   &  epochs& $20$ & $20$ & $20$ & $20$     \\  
 &  batch size  & $16$  & $32$  & $16$  & $16$     \\
  &  learning rate  & $3\mathrm{e}$-$5$    & $3\mathrm{e}$-$5$    &$5\mathrm{e}$-$5$    &$5\mathrm{e}$-$5$        \\  \hline

 \multirow{3}{*}{i2b2 2014}  &  epochs& $20$ & $20$ & $20$ & $20$     \\  
 &  batch size  & $16$  & $16$  & $32$  & $16$     \\
  &  learning rate  & $2\mathrm{e}$-$5$    & $2\mathrm{e}$-$5$    &$5\mathrm{e}$-$5$    &$3\mathrm{e}$-$5$        \\  \hline

  \end{tabular}
  \caption{Hyperparameter selection of all the models for each dataset}
  \label{tab:hype}
\end{table*}

\subsection{UmlsBERT Training}
 We initialize UmlsBERT with the pre-trained Bio\_ClinicalBERT model \citep{alsentzer-etal-2019-publicly}, and then we further pre-train it with the updated Masked LM task on MIMIC-III notes. 
 Afterwards, in order to perform the downstream tasks, we add a single linear layer on top of UmlsBERT and `fine-tuned' it to the task at hand, using either the associated embedding for each token or the embedding of the \textit{[CLS]} token. The same fine-tuning method is applied to all other models used for comparison. In order to keep the experiment controlled,  we use the same vocabulary and  WordPiece tokenization \citep{Wu2016GooglesNM} across all the models. WordPiece divides words not in the vocabulary into  frequent sub-words.
 
 Since our goal is to demonstrate the beneficial effect of incorporating domain knowledge in this study, we haven't experimented with a more complicated layer on top of UmlsBERT (e.g. the Bi-LSTM layer in \citep{Siarticle}). This is because our goal is to demonstrate that incorporating domain knowledge was beneficial for the performance of the model by showing that UmlsBert outperformed the other medical-based BERT models on a variety of  medical NLP tasks (Section \ref{results}). It should be noted that we chose the UMLS Metathesaurus in our process of  augmenting the UmlsBERT model for two reasons:  
 \begin{enumerate}
 \item We aim to create a clinical contextual embedding  model  that is capable  of  integrating  domain (medical)  knowledge.
 \item  The  UMLS Metathesaurus  is  a  compendium  of  many popular biomedical vocabularies (e.g. MeSH \citep{Mesh} and ICD-10 \citep{icd10}). By choosing to utilize the domain (medical) knowledge of UMLS, we actually incorporate domain knowledge from all major internationally standardized clinical terminologies.
 \end{enumerate}
 

 In the pre-training phase, UmlsBERT is trained for $1,000,000$ steps with a batch size of $64$, maximum sequence length of $128$ and learning rate of $5 \cdot 10^{-5}$. All  other hyper-parameters are kept to their default values. UmlsBERT is trained by using 2 nVidia V100 16GB GPU's with 128 GB of system RAM running Ubuntu 18.04.3 LTS.
 
\section{Results}
\label{results}

 \begin{table*}
 \centering
  \begin{tabular}{l c c c c c}
  \hline

Dataset &   & BERT\textsubscript{based}  &     BioBERT &   Bio\_ClinicalBERT &    UmlsBERT   \\ \hline
\multirow{3}{*}{MedNLI}  &  Test Ac.& 77.9 $\pm$  0.6  &  82.2 $\pm$0.5 &   81.2 $\pm$ 0.8 &    \textbf{83.0 $\pm$ 0.1} \\  
 &  Val. Ac.  & 79.0 $\pm$ 0.5   &    83.2 $\pm$ 0.8  &   83.4 $\pm$ 0.9 &    \textbf{84.5 $\pm$ 0.1} \\
  &  Run. time(sec)  & 308    &    307  & 269 &   305   \\ 
  &  \#parameters  & 108,312,579   &    108,312,579  & 108,312,579 & 108,346,371  \\ \hline
 
 \multirow{3}{*}{i2b2 2006}  &  Test F1 & 93.5 $\pm$ 1.4 &  93.3 $\pm$ 1.3 &  93.1 $\pm$ 1.3 &   \textbf{93.6 $\pm$ 0.5} \\  
 &  Val. F1  &  94.2 $\pm$ 0.6  &  93.8 $\pm$ 0.3 &   93.4 $\pm$ 0.2  &   \textbf{94.4 $\pm$ 0.2}   \\
  &  Run. time(sec)  & 12508   &   12807  &  12729 &   13167  \\ 
    &  \#parameters  & 108,322,576   &    108,322,576   & 108,322,576  &   108,356,368  \\ \hline

 \multirow{3}{*}{i2b2 2010}  &  Test F1& 85.2 $\pm$ 0.2   &   87.3 $\pm$ 0.1 &  87.7 $\pm$ 0.2 &   \textbf{88.6 $\pm$ 0.1} \\  
 &  Val. F1  & 83.4 $\pm$ 0.3 & 85.2 $\pm$ 0.6  &   86.2 $\pm$ 0.2   &  \textbf{ 87.7 $\pm$ 0.5}  \\
  &  Run. time(sec)  &  5325  &  5244   & 5279  &   5219   \\ 
  &  \#parameters  & 108,315,655   &   108,315,655  & 108,315,655 &   108,349,447  \\ \hline

 \multirow{3}{*}{i2b2 2012}  &  Test F1& 76.5 $\pm$ 0.2  &  77.8 $\pm$ 0.2 &   78.9 $\pm$ 0.1 &  \textbf{79.4 $\pm$ 0.1} \\  
 &  Val. F1  &  76.2 $\pm$ 0.7  & 78.1 $\pm$ 0.5  &  77.1 $\pm$ 0.4 & \textbf{ 78.3 $\pm$ 0.4}  \\
  &  Run. time(sec)  & 2413   &  2387  & 2403  & 2432     \\ 
  &  \#parameters  & 108,320,269   &   108,320,269  & 108,320,269 &  108,354,061   \\ \hline

 \multirow{3}{*}{i2b2 2014}  &  Test F1 &  \textbf{95.2 $\pm$ 0.1}   & 94.6 $\pm$ 0.2 &   94.3 $\pm$ 0.2 &   94.9 $\pm$ 0.1 \\  
 &  Val. F1  & \textbf{ 94.5 $\pm$ 0.4}  &   93.9 $\pm$ 0.5 &   93.0 $\pm$ 0.3 &   94.3 $\pm$ 0.5   \\
  &  Run. time(sec)  & 16738   &   17079  &  16643  &  16554   \\ 
  &  \#parameters  &108,343,339   &    108,343,339   & 108,343,339  &   108,377,131  \\ \hline

  \end{tabular}
  \caption{Results of mean $\pm$ standard deviation of five runs from each model   on the test and the validation test;  we use the acronym  Ac. for accuracy; average running time and number of parameters is also provided for each model. The number of parameters is different between datasets as we included the linear layers that were used on top of the Bert-based model for text and token classification; best values are \textbf{bolded};}
  \label{tab:res}
\end{table*}

In this section, we present the results of an empirical evaluation of the UmlBERT model. In particular, we provide a comparison between different  available BERT models to show the efficiency of our proposed model on different clinical NLP tasks. In addition, we provide the results of an ablation test to exam the effect of the semantic type embeddings on the performance of the model. Furthermore, we conduct a qualitative  analysis of the embedding of each model in order to illustrate how medical knowledge improves the quality of medical embeddings. Finally, we provide a visualized comparison of the embeddings of the words that are associated with semantic types between UmlsBERT and Bio\_ClinicalBert.

\subsection{Downstream Clinical NLP Tasks}

In this section, we report the results of the comparison of our proposed UmlsBERT model with the other BERT-based models on different downstream clinical NLP tasks described in Section \ref{data}.  All BERT-based models are implemented using the transformers library \citep{Wolf2019HuggingFacesTS} on PyTorch 0.4.1. All experiments are executed on a Tesla P100 16.3 GB GPU with 32G GB of system RAM on Ubuntu 18.04.3 LTS.

\subsubsection{Hyperparameter tuning}

In order to address the reproducibility concerns of the NLP community \citep{showyourwork},  we provide the search strategy and the bound for each hyperparameter as  follows: the batch size is set between 32 and 64, and the learning rate is chosen between the values  $2\mathrm{e}$-$5$, $3\mathrm{e}$-$5$ and $5\mathrm{e}$-$5$. For the clinical NER tasks, we take a similar approach to \citep{10.1093/bioinformatics/btz682} and set the number of training epochs to 20 to allow for maximal performance, except for MedNLI, for which we train the models on 3 and 4 epochs.

The best values are chosen based on validation set F1 values using the seqevals python framework for sequence labeling evaluation, due to the fact that it can provide an evaluation of a NER  task on entity-level\footnote{\url{https://github.com/chakki-works/seqeval}} for the i2b2 tasks and validation set accuracy, which is the standard metric for this task \footnote{\url{https://tinyurl.com/transformers-metrics}} 
for the MedNLI dataset. In the interest of providing a fair comparison, we also tune the hyperparameters of each model in order to demonstrate its best performance.  The final hyper-parameters selection of all the models for each dataset can be found in Table \ref{tab:hype}.

In order to achieve more robust results,  we run our model on five different (random) seeds (6809, 36275, 5317, 82958, 25368) and we provide the average  scores and standard deviation for the testing and the validation set. It should be noted that  BERT$_{base}$, BioBERT and  Bio\_ClinicalBERT have the exact same number of parameters as they use the same BERT-based architecture. However, because we introduce the semantic type embeddings into the UmlsBERT model, our model has an additional 33792 [the number of unique UMLS semantic types (44)  $\times$  transformer's hidden dimension(768)]  parameters \footnote{UmlsBert also contains an additional zero-filled vector, that we use as the semantic type vector   of the words that are not identified in UMLS, which was  not included in the calculation of the number of the parameters of the model.}. In Table \ref{tab:res}, we provide the number of parameters for each dataset where we include the linear layer on top of the BERT-based models for the text and token classification.

\begin{table*}
\centering
  \begin{tabular}{l c c c c c c c   }
  \hline
   & \multicolumn{2}{c}{\underline{ \textbf{ANATOMY}\hphantom{em}}} &  \multicolumn{2}{c}{\underline{ \textbf{DISORDER}\hphantom{ey999}}} & \multicolumn{2}{c}{\underline{\hphantom{etp}\textbf{GENERIC}\hphantom{etyem}}}    \\
      & feet & kidney     & mass  & bleeding  & school  & war    \\  
    &   &    &    &   &      \\ \hline

     & ft & liver & masses & bleed & college  & battle  \\
     BERT\textsubscript{based}  & foot  & lung   & massive & sweating  & university & conflict   \\\hline
   
    & foot  & liver  & masses & bleed  & college  & wartime   \\
    BioBERT & wrists & lung &  weight & strokes  & schooling  & battle   \\\hline
 
  & foot &  liver  & masses & bleed  & college & warfare    \\ 
 Bio\_ClinicalBERT & legs & lung &  weight & bloody   & university & wartime    \\\hline

   & foot &  \textbf{\underline{Ren}}   &\textbf{\underline{lump}}  &  bleed &college& warfare   \\
     UmlsBERT & \textbf{\underline{pedal}}&  liver  & masses   & \textbf{\underline{hem}}   & students & military   \\\hline

  \end{tabular}
  \caption{ The two nearest neighbors for six words in three semantic categories (two clinical and one generic). Note that only UmlsBERT finds word associations based on the CUIs of the UMLS Metathesaurus that have clinical meaning whereas in the generic category, there is no discernible discrepancies between the models. }
  \label{tab:1}
\end{table*}

\subsubsection{BERT-based model comparison}
\label{down_results}

The  mean and standard deviation (SD) of the scores for 
all the competing models on  different NLP tasks are reported in Table \ref{tab:res}. UmlsBERT achieves the best results in 4 out of the 5 tasks. It achieves the best F1 score in three i2b2 tasks (2006, 2010 and 2012) ($93.6\%$, $88.6 \%$ and $79.4 \%$) and the best accuracy on the MedNLI task ($83.0 \%$).

Because our model is initialized with Bio\_ClinicalBERT model and pre-trained on the MIMIC-III dataset, it is not surprising that it does not outperform the BERT model on  i2b2 2014  (The BERT\textsubscript{base} model achieved  $95.2 \%$ on i2b2 2014). This is probably due to the nature of the de-ID challenges which is described in detail in  \citep{alsentzer-etal-2019-publicly}. In summary, \textit{protected health information} (PHI) are replaced with  a sentinel `PHI' marker in the MIMIC dataset, but in the  de-ID challenge dataset (i2b2 2014), the PHI is replaced with different synthetic masks, and thus, the sentence structure that appears in BERT's training is not present at the down-stream task \citep{alsentzer-etal-2019-publicly}. However, even in this task, UmlsBERT achieves a better performance than the other biomedical BERT models.

These results confirm that augmenting contextual embedding through domain (biomedical) knowledge  is indeed  beneficial for the model's performance in a variety of biomedical down-stream tasks.

\subsubsection{Effect of semantic type embeddings}
In order to understand the effect that semantic type embeddings have on the model performance,  we  conduct an ablation test where the performance of two variations of the  UmlsBERT model are compared, where in one model the  semantic type embeddings are available to it, and in the other, they are not. 
The results of this comparison are listed in Table \ref{tab:res_abl}. 
We observe that for every dataset, UmlsBert achieves its best performance when semantic type embeddings are available. This experiment further confirms the positive effect of the semantic type embeddings on the performance of the UmlsBERT model.  
 \begin{table}[h!]
 \centering
  \begin{tabular}{l c c  c}
  \hline

Dataset &  & UmlsBERT$_{-ST}$   &    UmlsBERT   \\ \hline
MedNLI  &  Ac.& 82.3 $\pm$  0.2  &     83.0 $\pm$ 0.1 \\

i2b2 2006  &  F1 & 93.3 $\pm$ 0.7  & 93.6 $\pm$ 0.5  \\

i2b2 2010 &    F1& 88.3 $\pm$ 0.3&    88.6 $\pm$ 0.1 \\

i2b2 2012  &   F1& 79.1 $\pm$ 0.2&   79.4 $\pm$ 0.1 \\

i2b2 2014  &   F1 &   94.7 $\pm$ 0.1 &    94.9 $\pm$ 0.1 \\

  \end{tabular}
  \caption{Results of mean $\pm$ standard deviation of five runs for both variations of UmlsBERT on the test sets of all the datasets; In UmlsBERT$_{- ST}$,  the semantic type embeddings are not available.}
  \label{tab:res_abl}
\end{table}

\subsection{Qualitative Embedding Comparisons}
 Table \ref{tab:1} shows  the  nearest  neighbors  for  6  words   from  3  semantic categories  using  UmlsBERT, Bio\_ClinicalBERT, BioBERT and BERT. The first two categories (`ANATOMY' and `DISORDER') are chosen to demonstrate the ability of the models to identify similar words in a clinical context, and the third category (`GENERIC') is used to validate that the medical-focus BERT models  can  find meaningful  associations between words  in a general domain even if they are trained on  medical-domain text datasets. 
 
 This analysis demonstrates that augmenting the contextual embedding of UmlsBERT with Clinical Metathesaurus (UMLS) 
 information is indeed 
 beneficial for discovering associations between words with similar meanings in a clinical context.
 For instance, 
 only UmlsBERT discovers the connection between `kidney' and `ren' (from the latin word `renes', which means kidneys), between `mass' and `lump', between `bleeding' and `hem' (a commonly used term to refer to blood) and between `feet' and `pedal'(a term pertaining to the foot or feet in a medical context).
 
 These associations are the result of changing the nature of the Masked LM training phase of UmlsBERT to a multi-label scenario by connecting different words which share a common CUI in UMLS. In the previously mentioned examples, `kidney' and  `ren' have \textit{CUI:C0022646}; `mass' and `lump' have \textit{CUI:C0577559}; `bleeding' and `hem' have \textit{CUI:C0019080} and `feet' and `pedal' have \textit{CUI:C0016504}. 
 
 Finally, the results in the generic list of words indicate that the medical-focused BERT models did not trade their ability to find meaningful associations in a general domain in order to be more precise in a clinical context as there is no meaningful difference observed in the list of neighbour words that the four models identified.

\subsection{Semantic Type Embedding Visualization}


\begin{figure}[h]\centering
 \subfloat{\includegraphics[width=0.24\textwidth]{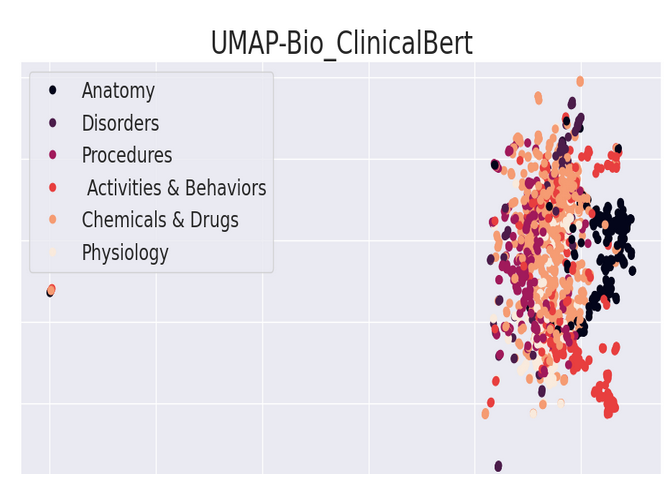}  \label{bio_umap} }
 \subfloat{\includegraphics[ width=0.23\textwidth]{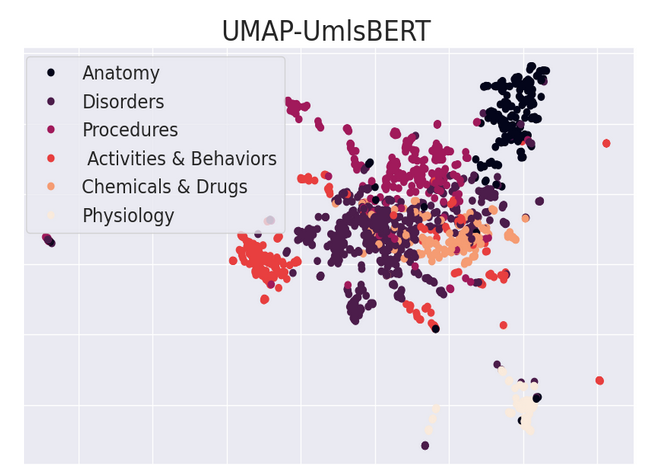} \label{umlsbert_umap}  }
 
 (a) \hphantom{emptyemp emp empt } (b)
 
  \caption{UMAP visualization of the clustering \textbf{(a)}  of the Bio\_ClinicalBert input embedding (word embedding)  \textbf{(b)} of the UmlsBert input embedding (word embedding + semantic type embedding).}
  \label{fig:clust}
\end{figure}

In order to demonstrate the effect of the semantic types 
on the 
input 
embeddings,   we present  in Figure \ref{fig:clust},  a UMAP dimensionality reduction \citep{umap} mapping comparison between Bio\_ClinicalBERT and UmlsBERT. We compare the input embedding 
of Bio\_ClinicalBERT with the input embedding 
of UmlsBERT for all the clinical terms that UMLS identified in the standard BERT vocabulary. It should be noted that in the graph, we group the medical terms by their semantic groups,  which are 
clusters that consist  of different semantic types. For example, the semantic types `Cell' and  `Body System' are  grouped in the semantic group `ANATOMY'.
It is evident that the clustering according to the semantic group that exists in the UmlsBERT embeddings (Figure \ref{umlsbert_umap}) cannot be found in the Bio\_ClinicalBERT embeddings (Figure \ref{bio_umap}). Thus,  we can conclude that more meaningful input embeddings can be provided to the model,  by augmenting the input layer of the BERT architecture with  the semantic type vectors, as they 
force the embeddings of  the words of the same semantic type to become more similar.

\section{Conclusion and Future Work}
\label{conlusion}
This paper presents UmlsBERT, a novel BERT-based architecture that incorporates domain (biomedical) knowledge in the pre-training process of a contextual word embeddings model. We demonstrate that  UmlsBERT can learn the association of different clinical terms with similar meaning in the UMLS Metathesaurus. UmlsBERT can also create more meaningful input embeddings by leveraging information from  the semantic type of each (biomedical) word. Finally, we confirm that these modifications can improve the model's performance as our UmlsBERT model outperforms other biomedical BERT models in various downstream tasks.

As for future work, we plan to address the limitations of this study including: \begin{enumerate*}[label=(\roman*)] \item Examining the effect of augmenting contextual embeddings with medical knowledge when more complicated layers are used atop of the output embedding of UmlsBERT. 
\item Exploring the UMLS  hierarchical associations between words that extend the concept connection that we investigated in this paper.  \item Testing our model in other datasets and biomedical tasks (e.g. relation extraction task  \cite{Krallinger2017OverviewOT}) to investigate further the strengths and weaknesses  of our model. 
\end{enumerate*}

\section{Acknowledgement}
We acknowledge the generous support from Microsoft AI for Health Program, MITACS Accelerate grant (\#IT19239), Semantic Health Inc., NSERC and Canada Research Chairs Program.

\section*{Ethical Considerations}

Contextual word embeddings models have achieved state-of-the-art results in many (clinical) NLP tasks such as NER or relation extraction \cite{Devlin2019BERTPO, 10.1093/bioinformatics/btz682}. These results suggest that  medical-based contextual word embeddings models, such as our model (UmlsBERT), can be a valuable tool for better processing and understanding the vast volume of health data that is  amassed at a rapid speed in health and biomedical domain. 

However, one of obstacles for adopting such a model in any system lies in the computing cost of pre-training. For example, our UmlsBERT model was trained for 10 days using 2 nVidia V100 16GB GPU's with 224 GB of system RAM running Ubuntu 18.04.3 LTS, and we acknowledge that investing these types of computational resources or even time is not a viable option for many research groups, let alone regular healthcare providers. This is the reason for making the UmlsBert  model publicly available, as we hope that the clinical NLP community can benefit from  using our model. In addition, UmlsBERT is the first contextual word embedding model, to the best of our knowledge, that integrated  structured medical-domain knowledge into its pre-training phase. Although this study demonstrates the beneficial effect of incorporating  structured biomedical domain knowledge in the pre-training phase of a contextual embedding model on the performance of the model, it is not a far-fetched hypothesis that similar pre-training strategy can be applied to incorporate structured domain-knowledge in different disciplines (e.g. environment, sciences, etc) to improve the performance of the model in the respective domain-specific down-stream tasks.

Finally, we believe that many research groups in the clinical NLP field could benefit from the use of our models by either using the contextual embeddings of our model or fine-tuning our model in specific down-stream tasks, for example, automatic encoding of diseases and procedures in electronic medical records. This automatic encoding model can significantly reduce time and cost in data extraction and reporting.  Success in such task will have huge impact in clinical practices and research since assigning correct codes for diseases and clinical procedures are important for making  care or operational decisions in healthcare.  

\bibliography{custom}
\bibliographystyle{acl_natbib}

\appendix



\end{document}